\documentclass{svproc}
\usepackage{url}
\usepackage{graphicx}
\usepackage{amsmath}
\usepackage{amssymb}
\usepackage[colorlinks]{hyperref}

\begin{document}
\mainmatter              
\title{Comparative Analysis of Parameterized Action Actor-Critic Reinforcement Learning Algorithms for Web Search Match Plan Generation}
\titlerunning{Parameterized Action Actor-Critic Reinforcement Learning Algorithms}  
%
\author{Ubayd Bapoo \and Clement N Nyirenda }
\authorrunning{Bapoo and Nyirenda} 
%
\tocauthor{Ivar Ekeland, Roger Temam, Jeffrey Dean, David Grove,
Craig Chambers, Kim B. Bruce, and Elisa Bertino}
\institute{University of the Western Cape, Bellville 7535, South Africa,\\
\email{2934126@myuwc.ac.za, cnyirenda@uwc.ac.za}
}

\maketitle              

\begin{abstract}
This study evaluates the performance of Soft Actor Critic (SAC), Greedy Actor Critic (GAC), and Truncated Quantile Critics (TQC) in high-dimensional decision-making tasks using fully observable environments. The focus is on parametrized action (PA) spaces, eliminating the need for recurrent networks, with benchmarks Platform-v0 and Goal-v0 testing discrete actions linked to continuous action-parameter spaces. Hyperparameter optimization was performed with Microsoft NNI, ensuring reproducibility by modifying the codebase for GAC and TQC. Results show that Parameterized Action Greedy Actor-Critic (PAGAC) outperformed other algorithms, achieving the fastest training times and highest returns across benchmarks, completing 5,000 episodes in 41:24 for the Platform game and 24:04 for the Robot Soccer Goal game. Its speed and stability provide clear advantages in complex action spaces. Compared to PASAC and PATQC, PAGAC demonstrated superior efficiency and reliability, making it ideal for tasks requiring rapid convergence and robust performance. Future work could explore hybrid strategies combining entropy-regularization with truncation-based methods to enhance stability and expand investigations into generalizability. 
\keywords{Match Plan, Parameterised Action, Soft Actor-Critic, Greedy Actor-Critic, Truncated Quantile Critics}
\end{abstract}
\section{Introduction}
Over the past two decades, text search engine technology has advanced significantly through innovations in index design, storage, and query evaluation (Zobel and Moffat, 2006). However, many contemporary techniques remain underrepresented in textbooks, creating resource gaps. The rapid growth of web documents linked to recognized keywords has made processing them all in short timeframes impractical (Luo et al., 2021), necessitating scalable and efficient search mechanisms. Search engines address this by organizing documents into ranked posting lists and employing match plans—sequential rules balancing result quality and response time—to process billions of pages efficiently (Rosset et al., 2020).

Despite these advances, challenges such as varied match rules, unpredictable data distribution, and rigid designs persist, requiring frequent customization. Luo et al. (2021) reframed match plan generation as a reinforcement learning (RL) problem, eliminating predefined inputs like stopping quotas. Their approach dynamically parameterizes state signals and match plans for adaptive development. Actions are represented in a hybrid parameterized space using runtime signals and static query features, with sparse, delayed rewards based on result quality and response time. To address these challenges, Luo et al. proposed Parameterized Action Soft Actor-Critic (PASAC), a deep RL algorithm combining discrete actions and continuous parameters for high-level decision-making. Inspired by Bester et al. (2021) and Haarnoja et al. (2018), PASAC integrates parameterized actions with SAC’s maximum entropy framework, enabling simultaneous reward optimization and entropy maximization. This enhances exploration while maintaining a stable actor-critic architecture.

While PASAC’s SAC introduces a novel approach to match plan generation, emerging Actor-Critic reinforcement learning methods offer strong alternatives. Greedy Actor-Critic (GAC) (Neumann et al., 2023) improves convergence speed and stability, outperforming SAC in some tasks. Truncated Quantile Critics (TQC) (Kuznetsov et al., 2020) mitigates overestimation bias through distributional representations, truncation, and ensembling. These algorithms present diverse strategies for balancing exploration and exploitation in high-dimensional continuous action spaces, necessitating comparison with SAC in parameterized action spaces. This study evaluates SAC, GAC, and TQC in two games over 5,000 episodes to assess policy learning and decision-making effectiveness. Key metrics include evaluation returns, training efficiency, convergence speed, and reward maximization. By comparing these algorithms, the study aims to provide insights into their performance, resilience, and applicability, advancing understanding of parameterized reinforcement learning in high-dimensional settings.

The structure of the rest of this paper is as follows: Section 2 provides an overview of match plan generation, while Section 3 introduces actor-critic reinforcement learning. In Section 4, we describe the Parameterized Actor-Critic Algorithms employed in this study. Sections 5 outlines the simulation setup and the ensuing results, respectively. Finally, Section 6 offers the study's conclusions.

\section{Match Plan Generation Overview}
Rosset et al. (2020) explain that search engines represent documents using fields—descriptions sourced from various inputs. Each field includes an inverted index where keywords are stored with posting lists, organized by document-level properties like static rank. During query processing, the system classifies the query into predefined types and generates a match plan to guide document scanning with specific criteria and stopping conditions, as illustrated in Figure 1. The merged documents are then subject to ranking and filtering to improve the quality of the search results. Match rules define the constraints a document must meet for ranking, while stopping criteria dictate when to end the index scan or proceed to the next rule. Together, they balance fast query responses with high-quality results. Complex or rare queries may require deeper matching strategies to uncover relevant documents. After executing the match plan, the selected candidates are refined using machine learning models to ensure optimal ranking and relevance.

\begin{figure}[h!]
    \centering
    \includegraphics[width=\textwidth]{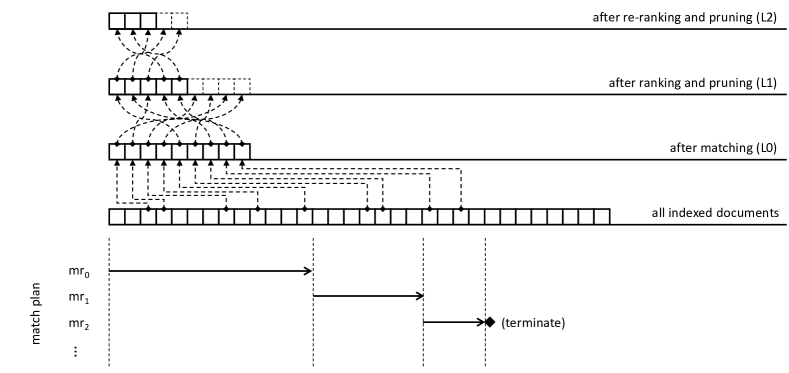}
    \caption{Illustration of the telescoping design used in Bing's retrieval mechanism.}
    \label{fig:label}
\end{figure}

\section{Overview on Actor-critic Reinforcement Learning}
Actor-critic methods are a class of algorithms in Reinforcement Learning (RL) (Sutton and Barto, 2018). Before delving into the details of actor-critic methods, it is important to first provide an overview of the fundamentals of RL. 

\subsection{RL Fundamentals}
RL is a machine learning subfield focused on teaching agents to act in environments to maximize cumulative rewards. Unlike supervised learning, which relies on labeled data, RL involves learning through interactions, enabling agents to explore and exploit their environment to discover optimal strategies (Sutton and Barto, 2018). An agent observes the current state of the environment \( s_t \) at time \( t \) and chooses an action \( a_t \) based on a policy \( \pi \), which is a mapping from states to actions:
\begin{equation}
\pi: S \to A,
\end{equation}
where \( S \) represents the state space, and \( A \) represents the action space. After the agent takes an action \( a_t \), the environment responds by transitioning to a new state \( s_{t+1} \) and providing a scalar reward \( r_t \). The agent's objective is to maximize the expected sum of rewards over time, typically represented by the return \( G_t \), which is defined as the discounted sum of future rewards:
\begin{equation}
G_t = \sum_{k=0}^{\infty} \gamma^k r_{t+k},
\end{equation}
where \( \gamma \) is the discount factor (with \( 0 \leq \gamma \leq 1 \)), representing the importance of future rewards relative to immediate rewards. RL has two primary components: the policy and the value function, which are described in more detail as follows:

\begin{enumerate}
\item \textit{Policy}: A strategy used by the agent to decide which action to take in a given state. It can be deterministic or stochastic. For a given state \( s_t \), the policy determines the probability of taking action \( a_t \):
\begin{equation}
   \pi(a_t | s_t) = P(a_t | s_t)
\end{equation}
   
\item \textit{Value Function}: A function that estimates the expected return from a given state (or state-action pair). The state-value function \( V(s) \) estimates the expected return starting from state \( s \) and following the policy \( \pi \):
\begin{equation}
   V^\pi(s) = \mathbb{E} [ G_t | s_t = s ]
\end{equation}
For an action-value function \( Q(s, a) \), the expected return is based on both the state and action:
\begin{equation}
   Q^\pi(s, a) = \mathbb{E} [ G_t | s_t = s, a_t = a ]
\end{equation}
\end{enumerate}

Value-based methods, such as Q-learning (Watkins and Dayan, 1992), SARSA (Rummery and Niranjan, 1994), and Deep Q-learning (Mnih et al., 2015), focus on estimating the value function and derive the policy indirectly by acting greedily with respect to this function. These methods have strengths, such as their ability to learn effectively in discrete action spaces, but struggle in environments with continuous action spaces, as finding the action that maximizes becomes computationally challenging. However, policy-based methods, such as REINFORCE (Williams, 1992), directly optimize the policy by using gradient ascent to maximize expected return. These methods excel in continuous action spaces and naturally handle stochastic policies, which are beneficial in exploration. However, they suffer from high variance in gradient estimates, leading to slow convergence and instability.

\subsection{Actor-critic Techniques}
Actor-critic methods are a class of RL algorithms that merge value-based and policy-based approaches, providing a robust framework for complex decision-making tasks. These techniques involve two components: the actor and the critic (Konda and Tsitsiklis, 1999; Sutton and Barto, 2018). The actor selects actions based on a policy, while the critic evaluates these actions by estimating the value function. This separation exploits the strengths of both approaches, making actor-critic methods particularly effective in continuous action spaces (Silver et al., 2014) and large state spaces (Mnih et al., 2016). Mathematically, the actor-critic framework is defined by two primary functions:

\begin {enumerate}
\item \textbf{The Actor (Policy)}: The policy function $\pi(a|s; \theta)$ defines the probability of selecting an action $a$ in state $s$, given parameters $\theta$. The goal of the actor is to adjust its parameters to maximize the expected return $J(\theta)$, which is the cumulative reward that the agent expects to obtain over time:
\begin{equation}
    J(\theta) = \mathbb{E}\left[\sum_{t=0}^{T} \gamma^t r_t \mid s_0. a_0, \dots \right].
\end{equation}

\item \textbf{The Critic (Value Function)}: The critic estimates the value function $V(s)$, which evaluates the expected return for being in state $s$ and following the current policy. The value function is updated based on the temporal difference (TD) error, $\delta_t$, which is the difference between the expected value and the actual reward:
\begin{equation}
    \delta_t = r_t + \gamma V(s_{t+1}) - V(s_t).
\end{equation}
The critic uses the TD error to adjust the value function parameters $\phi$, typically using gradient descent methods:
\begin{equation}
    \phi \leftarrow \phi + \alpha \delta_t \nabla_{\phi} V(s_t),
\end{equation}
where $\alpha$ is the learning rate.
\end{enumerate}

The actor updates the policy parameters $\theta$ by utilizing the advantage function $A(s_t, a_t)$, which measures the difference between the actual return and the estimated value of the state-action pair. The advantage function is defined as:
\begin{equation}
    A(s_t, a_t) = Q(s_t, a_t) - V(s_t),
\end{equation}
where $Q(s_t, a_t)$ represents the action-value function. The policy update rule for the actor is typically performed using the policy gradient method:
\begin{equation}
    \theta \leftarrow \theta + \beta \nabla_{\theta} \log \pi(a_t|s_t; \theta) A(s_t, a_t),
\end{equation}
where $\beta$ is the learning rate (Williams, 1992). In these methods, the actor-critic interaction is key for efficient learning. The critic guides the actor by indicating actions that yield higher rewards, while the actor refines the policy based on the critic's feedback.

\subsection{Emerging Deep Learning based Actor-Critic Techniques}
Over the past decade, numerous actor-critic extensions, primarily leveraging deep learning architectures, have been introduced. These advancements have significantly improved both the stability and performance of reinforcement learning models. Notable examples include Deep Deterministic Policy Gradient (DDPG) (Lillicrap et al., 2015), Advantage Actor-Critic (A2C) (Mnih, 2016), Proximal Policy Optimization (PPO) (Schulman et al., 2017), Twin Delayed Deep Deterministic Policy Gradient (TD3) (Fujimoto et al., 2018), Soft Actor-Critic (SAC) (Haarnoja et al., 2018), Truncated Quantile Critics (TQC) (Kuznetsov et al., 2020), and Greedy Actor-Critic (GAC) (Neumann et al., 2023). This subsection focuses on the three most recent methods in this list, which were implemented and evaluated in this study. These techniques are detailed as follows:
\begin{enumerate}
\item \textbf{Soft Actor-Critic (SAC)}: SAC maximizes an agent's interaction with its environment through soft policy iteration, evaluating and improving policies under the maximum entropy framework (Ding et al., 2022). SAC outperforms other RL algorithms in continuous action benchmarks, excelling in learning speed and resilience (Ding et al., 2022). It optimizes the maximum entropy objective, balancing exploration and exploitation (Haarnoja et al., 2018). The entropy target promotes broad exploration, discards unfeasible options, and allows policies to capture multiple near-optimal behaviors, assigning equal probability to equally appealing actions.
\item \textbf{Greedy Actor-Critic (GAC)}: Neumann et al. (2023) introduce a cross-entropy method for policy improvement, iteratively selecting the top percentile of actions based on their learned values. The process focuses on the most effective actions for each state-action pair. \( N \) actions are sampled using a proposal policy, sorted by magnitude, and the policy is updated to increase the likelihood of the top \( \lceil \rho N \rceil \) actions for \( \rho \in (0, 1) \). Neumann et al. (2023) highlight GreedyAC's advantages over Boltzmann greedification, showing it guarantees policy improvement in the original MDP and prevents policy collapse by incorporating entropy into the proposal policy rather than the final actor policy. This design allows the agent to explore potentially optimal actions without bias in decision-making.
\item \textbf{Truncated Quantile Critics (TQC)}:Foo et al. (2022) applied CPT Actor and TQC in DRL trading strategies. TQC, a quantile-based distributional RL approach, mitigates overestimation bias in off-policy algorithms by combining three concepts: (i) learning a critic’s distributional representation, (ii) pooling critics, and (iii) truncating pooled critics. It uses an actor-critic architecture, where the truncated critics estimate a risk-sensitive Q-function, guiding action selection and resulting in a stable policy. Kuznetsov et al. (2020) propose truncating continuous distributional quantile critics to reduce overestimation bias by selecting atoms with the highest values and calculating the Q-value from the remaining atoms. This method improves Q-value evaluation accuracy by adjusting the number of atoms, resolving inflated overestimation caused by high return variation.
\end{enumerate}

\section{Parameterised Action Actor-Critic Algorithms}
This section outlines the implementation of parameterised action (PA) actor-critic algorithms, evaluated in this study. It begins by introducing the PA concept and its relevance to match plan generation. Next, it describes the formulation of the match plan generation problem as a reinforcement learning (RL) task. Finally, it details the integration of PA with the actor-critic techniques discussed in Section 3, providing a generalized framework for implementation.

Masson et al. (2016) describe a parameterized action as discrete but accompanied by a real-valued vector serving as its parameter. This modeling approach introduces structure by differentiating between continuous actions. Each step requires the agent to select an action and execute it with specified parameters. For instance, consider a football-playing robot capable of kicking, passing, and running. Each action can be associated with a continuous parameter vector, such as kicking the ball with a specified force, passing to a precise location, or running at a particular speed. These actions are parameterized uniquely based on their respective requirements. Using reinforcement learning (RL) for match plan generation in search engines presents several challenges:  

\begin{enumerate}
    \item \textit{Hybrid Action Space:}  
    The environment for match plan creation combines discrete and continuous actions, challenging conventional RL algorithms that handle only one action type at a time.

    \item \textit{Complex Environment:}  
    Search engine settings are complex, with diverse queries and dynamic state variables. Queries require high-dimensional representations, and intermediate state signals span wide continuous value ranges, increasing complexity.

    \item \textit{Sparse Rewards:}  
    Sparse rewards hinder RL training, as early policies offer limited guidance, leading to infrequent positive feedback and slower convergence.
\end{enumerate}

Addressing these challenges requires advanced techniques capable of handling hybrid action spaces, managing high-dimensional data, and mitigating the limitations of sparse reward signals. The Parameterised Action (PA) Actor-Critic Framework handles hybrid action spaces, combining discrete and continuous actions. In the original SAC, an actor network estimates the mean and variance of a Gaussian distribution for continuous actions, while a critic network evaluates their Q values. PASAC extends this by using separate actor branches to generate both action types simultaneously. These branches share initial layers to encode state information but differ in configurations. PASAC uses a single critic network to estimate Q values for both types. Unlike traditional actor-critic methods, the PA framework integrates parameterised actions through significant modifications.

\begin{enumerate}
\item \textit{Dual Actor Networks}: The actor network generates discrete and continuous actions simultaneously, with shared state-encoding layers that diverge into specialized structures for each action type.
\item \textit{Unified Critic Network}: A single critic network evaluates the combined Q-values of discrete and continuous actions, maintaining a consistent hybrid action space representation during training.
\item \textit{State Representation Sharing}: Sharing parameters in the early actor network layers enables efficient state representation and flexibility in creating discrete and continuous actions.
\end{enumerate}

This extension of the actor-critic paradigm allows for successful decision-making in contexts where actions are parameterised, providing a formal answer to the issues presented by hybrid action spaces.

\section{Simulation Setup and Results}
This study evaluates the adaptability and efficacy of Soft Actor Critic (SAC), Greedy Actor Critic (GAC), and Truncated Quantile Critics (TQC) in high-dimensional decision-making with fully observable environments, removing the need for recurrent networks. The benchmarks, Platform-v0 and Goal-v0, test agents on discrete actions linked to continuous action-parameter spaces. Hyperparameter optimization was performed using Microsoft NNI, with the same codebase modified for GAC and TQC to ensure reproducibility. Ten tests were conducted to assess algorithm stability, efficiency, and performance, based on mean training incentives, final evaluation scores (averaged over 10 repetitions), and time to complete 5,000 episodes.

\subsection{Platform Game Description and Results}
The Platform domain features three actions: run, hop, and leap, each with a continuous parameter for horizontal displacement. To reach the target state, the agent must hop over opponents and leap across gaps. Contact with an opponent or falling through a gap results in failure. The 9-dimensional state space includes the agent's and local opponent's positions and velocities, along with platform characteristics such as length (Bester et al., 2021). Table \ref{tab:platform_game_results} presents the results for the Platform Game. PAGAC achieves the fastest training times in the Platform game, averaging 41:24 for the benchmark. It completes 5,000 episodes quicker than the other three algorithms while achieving the highest average evaluation return. With a maximum return of 0.1705, PAGAC consistently outperforms its counterparts.
\begin{table}[h!]
\centering
\caption{Performance Results for Platform Game }
\begin{tabular}{|l|l|l|}
\hline
\textbf{Algorithm} &\textbf{Average Evaluation Return} & \textbf{Average Time (hh:mm:ss)} \\ \hline
\textbf{PAGAC}     &\textbf{0.170}    &\textbf{0:41:24}           \\ \hline
PASAC              &0.143             &0:56:35           \\ \hline
PATQC              &0.145             &0:51:38           \\ \hline
\end{tabular}
\label{tab:platform_game_results}
\end{table}

\subsection{Robot Soccer Goal Game Description and Results}
The Robot Soccer Goal game simplifies RoboCup 2D, tasking an agent to score against a goalie intercepting the ball. The three parameterised actions—kick-to, shoot-goal-left, and shoot-goal-right—require kicking the ball, with the agent automatically approaching it until close enough to kick again. The state space includes 14 continuous features describing the positions, velocities, and orientations of the agent, goalie, and ball, along with the ball’s distance from the keeper and goal (Bester et al., 2021). Table \ref{tab:soccer_goal_results} presents the results for the Robot Soccer Goal Game. PAGAC achieves the fastest training time in the Goal game, averaging 24:04 for 5,000 episodes. It outperforms the other algorithms. PATQC slightly surpasses PASAC in average return, scoring -8.1552 compared to PASAC’s -8.2003.

\begin{table}[h!]
\centering
\caption{Performance Results for Soccer Goal Game }
\label{tab:soccer_goal_results}
\begin{tabular}{|l|l|l|}
\hline
\textbf{Algorithm} &\textbf{Average Evaluation Return} & \textbf{Average Time (hh:mm:ss)} \\ \hline
\textbf{PAGAC}     &-8.552              & \textbf{0:24:04}          \\ \hline
PASAC              &-8.200              & 0:26:55                   \\ \hline
PATQC              &\textbf{-8.155}     & 0:27:19                   \\ \hline
\end{tabular}
\end{table}

In general, these results show that PAGAC consistently achieved the fastest training times: 41:24 for Platform and 24:04 for Goal. PATQC followed with 51:38 for Platform and 27:19 for Goal, while PASAC was slowest at 56:35 and 26:55, respectively. Although PATQC slightly outperformed PASAC in returns, PAGAC excelled in speed and performance stability, proving its advantages in high-dimensional continuous action spaces. These findings underscore PAGAC’s efficiency for rapid convergence and reliable decision-making. Alongside PATQC, it provides efficient, robust solutions, reaffirming prior research.
\section{Conclusion and Future Works}
This work compared the soft actor critic (SAC) with the greedy actor critic (GAC) and the truncated quantile critic (TQC) in parametrizated action (PA) spaces. Using fully observable environments, it focused on optimizing parameterized action spaces without recurrent networks. Benchmarks Platform-v0 and Goal-v0 tested agents with discrete actions tied to continuous parameters, challenging their ability to handle complex decisions. The results revealed that parametrizated action greedy actor-critic (PAGAC) outperformed its counterparts. PAGAC consistently trained faster and provided stable, reliable decision-making, excelling in efficiency and performance. Truncated Quantile Critics (TQC) and Soft Actor Critic (SAC) delivered competitive outcomes but lagged in efficiency, with SAC being the slowest. Future research could explore hybrid strategies combining entropy-regularization methods (Haarnoja et al., 2018) with truncation-based approaches (Kuznetsov et al., 2021) to enhance stability and performance. Expanding the scope of the investigation may also provide valuable insights into the generalizability of the methodologies.
%
%

\end{document}